\documentclass[sigconf,screen,nonacm]{acmart}

\usepackage{booktabs}
\usepackage{graphicx}
\usepackage{amsmath}
\usepackage{multirow}
\usepackage{array}
\usepackage{xcolor}

\newcommand{\bk}{\discretionary{}{}{}}
\newcommand{\fpath}[1]{\texttt{#1}}

\begin{document}

\title{Fidelity Preference, Not Demographic Preference: A Pixel-Level
Attribute-Sensitivity Audit of Image Aesthetic/Preference Scorers}

\author{Mingyang Xu}
\authornote{Corresponding author.}
\affiliation{%
  \institution{Peking University}
  \city{Beijing}
  \country{China}
}
\email{2501220225@stu.pku.edu.cn}

\renewcommand{\shortauthors}{Xu}

\begin{abstract}
Text-to-image systems rely on learned aesthetic scorers to filter training data
and guide generation, but whether these scores encode demographic attributes as
``objective quality'' remains unclear. We audit 4 mainstream scorers
(LAION-Aesthetics, PickScore, ImageReward, HPSv2) using pixel-level
within-image interventions on skin tone and body type, across synthetic and
real data. Our core finding is that along a continuous skin-lightness axis the
dominant effect is not a directional demographic preference but a
\textbf{fidelity preference}: the unaltered image scores near-highest and
graded perturbation in either direction is penalized (inverted-U response).
Placebo arms establish that this penalty is not an artifact of the skin
operator---applying the same CIELAB $L^*$ shift to the \emph{non-skin} region
yields a penalty statistically indistinguishable in magnitude (LAION-Aes ratio
0.81, CI [0.54,1.19]; PickScore 1.22, CI [0.91,1.62])---while also bounding the
claim: the penalty is operator-dependent in size and holds on all tested
operators only for LAION-Aes. Critically, audits on synthetic images alone are
misleading: LAION-Aes shows strong preference for darker skin on SDXL faces
(asymmetry $-0.211$, best-of-$n$ amplification $-26.1^\circ$ ITA), but on 1470
real faces the asymmetry \textbf{flips sign} to $+0.032$ ($6.5\times$ smaller,
both arms FDR-significant) and the amplification \textbf{collapses to
non-significance} ($+0.2^\circ$). Across all four scorers the synthetic reading
fails to transfer---inverting for LAION-Aes and HPSv2, merely attenuating for
PickScore---so we frame the claim as non-transfer of synthetic effects rather
than universal reversal. We contribute: (1) a reproducible benchmark with
intervention-artifact covariate control and synthetic/real cross-validation;
(2) an attribute-auditability criterion defining when pixel-level causal
isolation yields valid bias measurements (skin tone: yes; body type: no, due to
unavoidable deformation artifacts). Population-stratified analysis shows
fidelity-penalty asymmetry is not robust across demographic groups after FDR
correction (only HPSv2 survives). Our findings demonstrate that naive synthetic
audits systematically misjudge both direction and magnitude of bias, and only
within-image causal isolation on real data can distinguish true demographic
bias from fidelity preference.
\end{abstract}



\maketitle

\section{Introduction}

Text-to-image (T2I) systems rely on learned aesthetic/preference scorers to
filter training data, perform best-of-$n$ sampling, and guide RLHF alignment.
Four mainstream scorers---LAION-Aesthetics V2, PickScore, ImageReward, and
HPSv2---assign scalar ``quality'' scores that directly shape generative model
behavior. A critical fairness question is whether these scores encode
demographic attributes (skin tone, body type) as objective quality, thereby
systematically filtering or reranking certain populations.

Existing audits mostly use cross-image comparisons on synthetic data, which
confounds attribute changes with composition, lighting, and rendering
artifacts. \textbf{We find that this approach misjudges both the direction and
magnitude of bias.} Using pixel-level within-image causal isolation on 210 real
faces and 54 synthetic counterfactuals, we show that what appears as
``monotonic preference for darker skin'' ($-26^\circ$ ITA amplification) on
synthetic images \textbf{collapses to a near-symmetric inverted-U fidelity
preference} on real faces ($+0.2^\circ$ amplification, not significant). The
scorer penalizes \emph{any} directional perturbation from the original pixel
values---not a demographic preference, but a pixel-level fidelity preference.

This reading is independently corroborated from a direction our method cannot
see. Concurrent work auditing the LAION-Aesthetics Predictor at corpus scale,
together with a trace ethnography of its construction, concludes that it ``could
be more accurately described as a measure of photorealism''~\cite{taylor2026gaze}.
That conclusion comes from between-image distributional evidence and provenance
analysis; ours comes from within-image graded intervention. Neither implies the
other, which is why their agreement matters (Section~\ref{sec:related}).

This distinction matters for fairness. A fidelity preference indirectly rewards
training-distribution modes and penalizes long-tail appearances. Whether this
penalty falls asymmetrically across demographic groups is an empirical question
we address via stratified analysis (Section~\ref{sec:rigor}). Our contribution
is threefold: (1) \textbf{a reusable audit benchmark} with
intervention-artifact covariate calibration and synthetic/real
cross-validation; (2) \textbf{distinguishing fidelity preference from
demographic bias} via within-image dose-response curves; (3) \textbf{an
attribute-auditability diagnostic} defining when pixel-level causal isolation is
valid (skin tone: yes; body type: no).

\section{Related Work and Positioning}
\label{sec:related}

\paragraph{Scorer-bias critiques.}
Concerns that aesthetic/preference scorers encode social preferences predate
this work, raised both qualitatively and---more recently---at corpus scale (see
\cite{taylor2026gaze} below). The same failure mode has been documented outside
vision: Zhang et al.~\cite{zhang2026genre} show genre-induced shortcut learning
in music aesthetics scoring, where the model uses genre features as a proxy for
quality, and \cite{guo2025universal} argue that image reward models penalize
deliberately anti-aesthetic outputs even when these match the user's explicit
prompt. What no prior audit has done is causally separate an attribute's effect
from compositional confounds \emph{within a single image}, or distinguish
``demographic-direction preference'' from ``fidelity preference.''

The 4 scorers audited here represent different training signals:
LAION-Aesthetics is used to filter the SD training set from
LAION-5B~\cite{schuhmann2022laion}; PickScore~\cite{kirstain2023pick},
ImageReward~\cite{xu2023imagereward}, and HPSv2~\cite{wu2023hps} are reward
models built on human preferences. Among the four, LAION-Aes, PickScore, and
HPSv2 share a CLIP~\cite{radford2021clip} visual backbone, while ImageReward
uses a BLIP architecture, covering two mainstream families of vision-language
encoders. Demographic-stereotype amplification on the T2I generation side has
been systematically documented (Bianchi et al.~\cite{bianchi2023easily},
Luccioni et al.'s Stable Bias~\cite{luccioni2023stable}); we focus on the
complementary scoring side---whether the scorer itself encodes demographic
attributes as quality.

\paragraph{Auditing the same scorer at corpus scale (nearest prior work, same
object of study).}
Taylor et al.~\cite{taylor2026gaze} audit the LAION-Aesthetics Predictor
itself---one of the four scorers we audit---by scoring three large corpora (the
$\sim$1.2B-image LAION-Aesthetics Dataset, $\sim$249k Metropolitan Museum of Art
images, $\sim$81k WikiArt images) and pairing this with a trace ethnography of
how the predictor was built. They report that filtering at the widely used 6.5
threshold disproportionately retains images whose captions mention women while
filtering out those mentioning men or LGBTQ+ people, that western and Japanese
artists are rated most highly, and---the finding closest to ours---that the
predictor ``could be more accurately described as a measure of photorealism''
rather than of aesthetic quality, an origin they trace to training images that
are predominantly photographs rated largely by English-speaking photographers
and western AI enthusiasts.

\paragraph{This convergence is the strongest external support our mechanism
claim has, and it arrives by an entirely different route.}
\cite{taylor2026gaze} reach ``LAP measures photorealism'' from
\emph{between-image} distributional evidence over corpora plus provenance
analysis; we reach a compatible conclusion---that the dominant response on the
skin-lightness axis is a penalty on \emph{departure from the original pixels}
rather than a preference for a demographic direction
(Section~\ref{sec:skin})---from \emph{within-image} graded intervention on 210
real faces. Neither result implies the other. A corpus-level association between
photorealism and high scores does not by itself predict the shape of the
response to a controlled perturbation of one image; conversely, our inverted-U
says nothing about which corpora get filtered in. That two methods with disjoint
failure modes agree on ``this scorer is tracking rendering fidelity, not the
depicted subject's attributes'' is what makes the mechanism credible rather than
an artifact of either design.

\paragraph{We are careful about the boundary.}
Three of \cite{taylor2026gaze}'s findings are outside what our design can speak
to, and we do not claim to corroborate or contest them: caption-mediated gender
and LGBTQ+ disparities (we intervene on pixels, never on captions, and use
neutral prompts throughout), cultural and art-historical ranking (our stimuli
are face and body photographs, not artworks), and the provenance claims from
trace ethnography (a qualitative method we do not employ). Conversely, our
synthetic-versus-real divergence (Section~\ref{sec:scale}) identifies a hazard
their design does not encounter: because they score \emph{existing} images
rather than \emph{generated counterfactuals}, they are not exposed to the
rendering confound that makes synthetic-only audits misjudge both direction and
magnitude. The two contributions are complementary in the strict sense---same
scorer, non-overlapping evidence, and each covering a validity threat the other
cannot reach.

\paragraph{Scorer-bias audits (nearest prior work on demographic attributes).}
The closest work to ours on the demographic axis is Abdel Magid et
al.~\cite{abdelmagid2026bias} (CVPR 2026), a large-scale audit of demographic
bias in T2I reward models. It reports that reward models encode demographic
biases as quality: reward-guided optimization disproportionately sexualizes
female subjects (PickScore raises the NSFW rate by 19\% for female vs.\ 7\% for
male subjects), collapses demographic diversity (ImageReward and HPS reclassify
$>$80\% of Black subjects as White after optimization), and produces consistent
categorical group rankings (HPS and ImageReward rank White subjects highest).
\textbf{We are complementary, not corrective, and we are careful about which
claim we engage.} \cite{abdelmagid2026bias} operates at the level of
\emph{categorical} demographic identity, using matched counterfactual image sets
(CausalFace, SocialCounterfactuals, PAIRS) in which ``identity attributes vary
while other semantic and contextual factors are held constant,'' and it measures
score disparities \emph{between} those sets. We operate at the level of a
\emph{single continuous} attribute axis (CIELAB $L^*$ skin lightness) and measure
the \emph{response shape} to graded within-image perturbation of that axis.
These are different targets: a $\pm 15$ $L^*$ shift does not reproduce a change
of categorical racial presentation, which also involves hair texture, facial
morphology, and other cues. \textbf{Our results therefore neither confirm nor
refute \cite{abdelmagid2026bias}'s categorical findings}, and we make no claim
that they do.

Where the two studies do meet is instructive, and it runs in
\cite{abdelmagid2026bias}'s favor rather than against it.
\cite{abdelmagid2026bias} explicitly considered and rejected skin tone as a
demographic proxy, on the grounds that ``skin tone estimates were highly
sensitive to generative model lighting conditions and shading'' and that ``skin
tone alone ignores other attributes such as hair texture.'' Our synthetic/real
divergence (Sections~\ref{sec:skin}--\ref{sec:scale}) is, in effect, an
independent quantification of exactly that concern: on SDXL-generated faces the
apparent skin-lightness preference is large (asymmetry $-0.211$, best-of-$n$
drift $-26.1^\circ$ ITA), and on real photographs the asymmetry flips sign to
$+0.032$ while the drift collapses to a non-significant $+0.2^\circ$. The
synthetic signal is a rendering confound, which is what
\cite{abdelmagid2026bias} anticipated qualitatively and what we measure. Our
contribution relative to \cite{abdelmagid2026bias} is thus methodological rather
than contradictory: pixel-level within-image isolation on \emph{real} data,
intervention-artifact covariate calibration, and an explicit criterion
(Section~\ref{sec:body}) for when an attribute admits such isolation at all.

\paragraph{Counterfactual fairness (methodological level).}
Counterfactual fairness~\cite{kusner2017counterfactual} requires ``change only
the protected attribute, hold everything else fixed.'' In T2I, swapping
attribute words in the prompt cannot achieve ``everything else fixed''---
composition, lighting, and pose all change along with it (the confounding
revealed in Section~\ref{sec:bench}). We use \textbf{pixel-level CIELAB
lightness intervention} (skin axis) and \textbf{torso geometric warp} (body
axis) to achieve ``everything else held fixed as much as possible, pixel by
pixel,'' which is the methodological core.

\paragraph{ITA$^\circ$ objective skin-tone scale / silhouette width
(measurement level).}
$\mathrm{ITA}^\circ = \arctan((L^*-50)/b^*)\cdot 180/\pi$ is the
dermatological-standard continuous scale of skin
lightness~\cite{chardon1991skin}; the body axis uses the relative width of the
torso band of the foreground silhouette, \texttt{width\_ratio}, as a continuous
objective measure. Both quantify ``the image's actual attribute value'' into a
continuous dose, rather than relying solely on prompt intent.

\begin{table*}[t]
\centering
\caption{Positioning of this work relative to the two nearest prior audits.}
\label{tab:positioning}
\footnotesize
\begin{tabular}{@{}p{0.13\textwidth}p{0.235\textwidth}p{0.235\textwidth}p{0.335\textwidth}@{}}
\toprule
\textbf{Dimension} & \textbf{Corpus/provenance audits
(\cite{taylor2026gaze})} & \textbf{Categorical counterfactual audits
(\cite{abdelmagid2026bias})} & \textbf{This work} \\
\midrule
Unit of analysis & Corpus of existing images & Matched counterfactual image
\emph{sets} & \textbf{Single image, graded perturbation} \\
\addlinespace[2pt]
Attribute treatment & Caption keywords, artist/culture metadata & Categorical
identity (race, gender) & \textbf{One continuous axis (CIELAB $L^*$), measured
per image via ITA$^\circ$} \\
\addlinespace[2pt]
Confound control & Large-$n$ distributional contrast & Attributes vary, context
held constant across sets & \textbf{Within-image pixel/geometric intervention +
intervention-artifact covariate calibration} \\
\addlinespace[2pt]
Mechanism evidence & Provenance (trace ethnography) $\rightarrow$ ``measures
photorealism'' & Score disparity between groups & \textbf{Response \emph{shape}
to dose (inverted U) + placebo operators} \\
\addlinespace[2pt]
Synthetic-vs-real & Real images only & Synthetic counterfactual sets &
\textbf{Both arms side by side}, showing synthetic effects do not transfer \\
\addlinespace[2pt]
Downstream & Filtering rates at a score threshold & Reward-guided optimization
outcomes & best-of-$n$ amplification curve + random null calibration \\
\addlinespace[2pt]
Auditability of the axis & Not addressed & Skin tone rejected as too confounded
& \textbf{Explicit criterion for when an axis admits pixel-level isolation
(Section~\ref{sec:body})} \\
\bottomrule
\end{tabular}
\end{table*}

Read across the row ``mechanism evidence'': \cite{taylor2026gaze} infer what the
scorer measures from \emph{where it came from}, \cite{abdelmagid2026bias} from
\emph{whom it scores differently}, and we from \emph{how it responds when one
thing changes}. The three are mutually reinforcing precisely because none of
them can substitute for the others.

\section{The Audit Benchmark: Three-Pillar Protocol + Dual Arms + Four Scorers}
\label{sec:bench}

We package the full audit into a reproducible benchmark \texttt{audit\_bench}
(config-driven, one-click pipeline). It is built on three methodological pillars.

\paragraph{Pillar 1: within-image intervention (causal isolation).}
Rather than comparing different images, we take the same base image and push a
single attribute while freezing the remaining pixels:
\begin{itemize}
\item \textbf{Skin axis}: shift the $L^*$ channel of CIELAB in the skin region by
$\Delta l \in \{-15, -10, -5, 0, +5, +10, +15\}$ (feathered blending at skin
mask boundary), composition unchanged pixel by pixel
(\fpath{interventions/\bk pixel\_ita.py}). Real arm: 210 base images from FairFace
validation set (1470 total images $=210\times7$ doses). Synthetic arm: 54 SDXL
counterfactuals (3 scenes $\times$ 3 seeds $\times$ 6 Fitzpatrick levels, 378
total images $=54\times7$ doses).
\item \textbf{Body axis}: anisotropically scale the width of the torso band by a
factor $\gamma \in \{0.8, 0.87, 0.93, 1.0, 1.07, 1.13, 1.2\}$ (Hann-window
smooth decay to 0 at head/legs), keeping background/head as fixed as possible
(\fpath{interventions/\bk body\_warp.py}). Synthetic arm: 54 base images (378 total
images $=54\times7$ doses).
\end{itemize}

\paragraph{Pillar 2: intervention-artifact covariate calibration.}
A within-image intervention may itself introduce artifacts rather than attribute
semantics: an $L^*$ shift may cause gamut over-clipping, and a geometric warp
introduces resampling deformation. Each intervention is paired with an artifact
measure used as a covariate in a discriminative regression: \texttt{clip\_frac}
(fraction of skin pixels clipped out of gamut) for the skin axis, and
\texttt{warp\_artifact} (max horizontal resampling displacement / image width)
for the body axis. The discriminative regression
$\mathrm{score} = \alpha_{\mathrm{base}} + \beta\cdot\mathrm{attribute} +
\gamma\cdot\mathrm{artifact}$ separates the attribute effect $\beta$ from the
artifact effect $\gamma$ in one shot.

\paragraph{Pillar 3: best-of-$n$ null calibration.}
best-of-$n$ selects the argmax from a candidate pool; even if the within-image
bias is tiny, argmax as a nonlinear amplifier accumulates a stably-directed bias
into a systematic drift. But argmax itself + an asymmetric attribute
distribution in the pool also mechanically produce drift. We subtract this off
using a random-selection baseline: net bias = argmax drift $-$ random drift
(paired bootstrap, 2000 resamples). We use $n=4$ candidates per selection.
Random should be $\approx 0$ (a validity check); significant net drift indicates
amplification beyond mechanical drift.

\paragraph{Core analysis tool: the fidelity curve.}
If the within-image dose-response is fit with a linear regression, the slope's
sign depends on which side of the response curve the dose-sampling interval falls
---unreliable (the fact that the 4 scorers' real-face linear $\beta$ have mixed
signs is evidence of this). The true mechanism is revealed by the
\textbf{within-base paired difference referenced to the original image}:
$\Delta_i(\mathrm{dose}) = \mathrm{score}_i(\mathrm{dose}) -
\mathrm{score}_i(\mathrm{original})$. If $\Delta$ is significantly negative on
both sides $\Rightarrow$ the original is the peak = inverted U =
\textbf{fidelity preference}. Two readings: the fidelity index (magnitude of
penalty on both sides) and the asymmetry (residual directional preference,
$\approx 0$ for pure fidelity).

\paragraph{Dual arms (parallel, not primary/secondary).}
The synthetic arm (SDXL~\cite{podell2024sdxl} counterfactuals, high
comparability but with rendering artifacts) and the real arm
(FairFace~\cite{karkkainen2021fairface} real faces, high ecological validity)
cross-validate, which is what allows us to distinguish ``an intrinsic property of
the scorer'' from a ``synthetic-rendering artifact.'' \textbf{The most important
lesson of this paper comes precisely from the failure of synthetic-arm effects to
transfer to the real arm (Sections~\ref{sec:skin},~\ref{sec:scale})---the real
arm is not a robustness appendix but a parallel main arm that overturns the
synthetic-arm conclusion.}

\paragraph{Four scorers (2 architecture families).}
LAION-Aes (CLIP-MLP regression), PickScore (CLIP-H contrastive), HPSv2 (CLIP-H
contrastive fine-tuned), ImageReward (BLIP cross-attention regression).
Conditional models uniformly use the neutral prompt ``a photo,'' measuring only
image-content preference and stripping out text-alignment.

\section{Skin-Tone Axis: Directional Reversal Between Synthetic and Real}
\label{sec:skin}

\paragraph{Cross-image reading is unreliable (synthetic, $n=54$).}
Fixing the identity skeleton and only swapping the Fitzpatrick I--VI skin-tone
slot, 3 scenes $\times$ 3 seeds $\times$ 6 levels. Reading the cross-image slope
of ``score $\sim$ ITA$^\circ$'' directly, both scorers' CIs straddle zero and are
not significant. Cross-image audits have neither detection power nor a
trustworthy direction---this is the motivation for all subsequent within-image
analysis, not a conclusion.

\begin{table*}[t]
\centering
\caption{Within-image intervention, \textbf{synthetic} faces ($n=54$ base
images $\times$ 7 $\Delta l$, $|\Delta l|\le 15$, \texttt{clip\_frac} constant
0). Fidelity curve referenced to the original ($\Delta l=0$).}
\label{tab:skin-synth}
\footnotesize
\begin{tabular}{lccc c}
\toprule
\textbf{Scorer} & \textbf{Fidelity index [95\% CI]} & \textbf{Asymmetry [95\%
CI]} & \textbf{Inv.\ U} & \textbf{best-of-$n$ net drift ($n{=}4$)} \\
\midrule
LAION-Aes & $-0.005$ [$-0.037$, $+0.028$] & $\mathbf{-0.211}$ [$-0.269$,
$-0.153$] \checkmark & No & $\mathbf{-26.1^\circ}$ [$-32.5$, $-19.1$]
\checkmark \\
PickScore & $+0.256$ [$+0.222$, $+0.291$] \checkmark & $-0.051$ [$-0.102$,
$+0.004$] & Yes & $-5.2^\circ$ [$-9.6$, $-1.1$] \checkmark \\
HPSv2 & $+0.0013$ [$+0.0002$, $+0.0023$] \checkmark & $-0.0027$ [$-0.0041$,
$-0.0014$] \checkmark & No & $-14.3^\circ$ [$-21.2$, $-7.3$] \checkmark \\
ImageReward & $-0.019$ [$-0.037$, $-0.002$] $\dagger$ & $+0.014$ [$-0.021$,
$+0.052$] & No & $-0.2^\circ$ [$-9.4$, $+8.7$] $\times$ \\
\bottomrule
\end{tabular}
\end{table*}

\begin{table*}[t]
\centering
\caption{Within-image intervention, \textbf{real} faces (FairFace, $n=210$ base
images $\times$ 7 $\Delta l$ = 1470 total images). Same intervention, same
analysis.}
\label{tab:skin-real}
\footnotesize
\begin{tabular}{lccc c}
\toprule
\textbf{Scorer} & \textbf{Fidelity index [95\% CI]} & \textbf{Asymmetry [95\%
CI]} & \textbf{Inv.\ U} & \textbf{best-of-$n$ net drift ($n{=}4$)} \\
\midrule
LAION-Aes & $\mathbf{+0.074}$ [$+0.057$, $+0.090$] \checkmark & $+0.032$
[$+0.005$, $+0.059$] \checkmark & \textbf{Yes} & $+0.2^\circ$ [$-2.3$, $+2.6$]
$\times$ \\
PickScore & $+0.088$ [$+0.070$, $+0.107$] \checkmark & $-0.008$ [$-0.031$,
$+0.015$] & \textbf{Yes} & $\mathbf{-3.2^\circ}$ [$-5.5$, $-0.9$] \checkmark \\
HPSv2 & $-0.0008$ [$-0.0014$, $-0.0003$] \checkmark & $+0.0022$ [$+0.0014$,
$+0.0029$] \checkmark & No & $+4.3^\circ$ [$+1.7$, $+7.1$] \checkmark \\
ImageReward & $+0.012$ [$-0.0001$, $+0.025$] & $+0.016$ [$-0.005$, $+0.038$] &
No & $-1.3^\circ$ [$-3.8$, $+1.2$] $\times$ \\
\bottomrule
\end{tabular}
\end{table*}

\begin{table*}[t]
\centering
\caption{How general is the divergence? On a strict criterion (\emph{both} arms
FDR-significant \emph{and} opposite in sign) only 4 of the 12 scorer $\times$
reading cells qualify.}
\label{tab:divergence}
\footnotesize
\begin{tabular}{@{}p{0.10\textwidth}p{0.27\textwidth}p{0.31\textwidth}p{0.28\textwidth}@{}}
\toprule
\textbf{Scorer} & \textbf{FI: synth $\rightarrow$ real} & \textbf{Asymmetry:
synth $\rightarrow$ real} & \textbf{Amplification: synth $\rightarrow$ real} \\
\midrule
LAION-Aes & $-0.005$ (ns) $\rightarrow$ $+0.074$ (sig) & $\mathbf{-0.211
\rightarrow +0.032}$, both sig: \textbf{reversal} & $-26.1^\circ$ (sig)
$\rightarrow$ $+0.2^\circ$ (ns): collapse \\
\addlinespace[2pt]
PickScore & $+0.256 \rightarrow +0.088$, both sig, same sign & $-0.051$ (ns)
$\rightarrow$ $-0.008$ (ns) & $-5.2^\circ \rightarrow -3.2^\circ$, both sig,
same sign \\
\addlinespace[2pt]
HPSv2 & $+0.0013 \rightarrow -0.0008$, both sig: reversal & $-0.0027
\rightarrow +0.0022$, both sig: reversal & $-14.3^\circ \rightarrow
+4.3^\circ$, both sig: reversal \\
\addlinespace[2pt]
ImageReward & $-0.019$ (sig) $\rightarrow$ $+0.012$ (ns): collapse & $+0.014$
(ns) $\rightarrow$ $+0.016$ (ns) & $-0.2^\circ$ (ns) $\rightarrow$
$-1.3^\circ$ (ns) \\
\bottomrule
\end{tabular}
\end{table*}

In the fidelity-index and asymmetry columns of Table~\ref{tab:skin-synth},
\checkmark~= FDR-significant in the 24-test FI/asymmetry grid (BH $\alpha=0.05$)
and $\dagger$~= raw-significant only (raw $p=0.034$, fdr $p=0.054$). In the drift
column, \checkmark/$\times$ = significant/not in the separate 36-test headline
grid (Section~\ref{sec:rigor}); the two grids are different measurement
conventions and are corrected independently.

\begin{figure}[t]
\centering
\includegraphics[width=\linewidth]{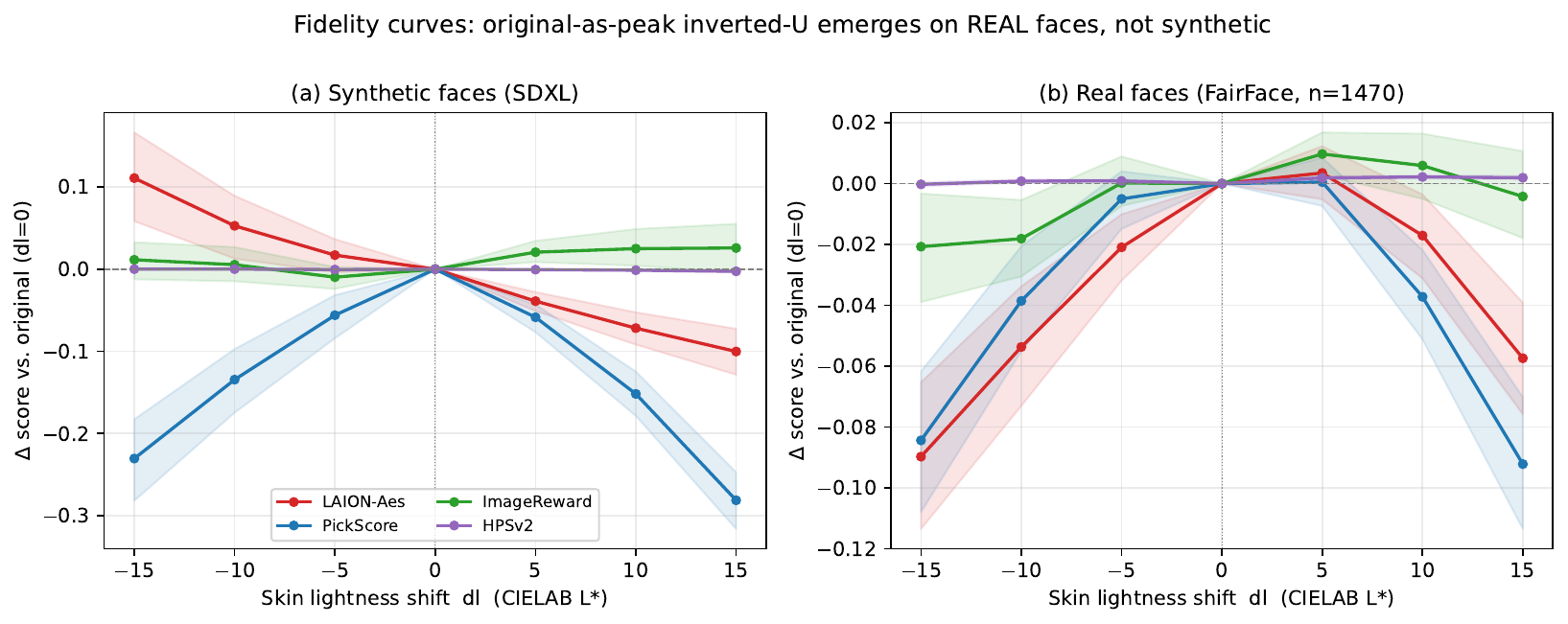}
\caption{Fidelity curves showing within-base score change ($\Delta$ score) as a
function of skin lightness intervention ($\Delta l$). (a) Synthetic faces show
LAION-Aes monotonic preference for darker skin (asymmetry $-0.21$). (b) Real faces
show symmetric inverted-U for LAION-Aes and PickScore, demonstrating fidelity
preference.}
\label{fig:fidelity}
\end{figure}

On synthetic faces LAION-Aes shows a \textbf{monotonic preference for darker
skin} (asymmetry $-0.21$, $\Delta l<0$ adds points, $\Delta l>0$ subtracts),
amplified by best-of-$n$ to $-26^\circ$ ITA (about one Fitzpatrick level). With
only the synthetic arm, this would be the strong conclusion ``LAION
systematically prefers darker skin, with severe downstream amplification.'' See
Figure~\ref{fig:fidelity}a for the full fidelity curves.

Markers in Table~\ref{tab:skin-real} are as above (FI/asymmetry columns: 24-test
grid; drift column: 36-test grid). Note LAION-Aes's residual asymmetry survives
FDR in the 24-test grid (raw $p=0.022$, fdr $p=0.040$): small, but not
attributable to noise.

\begin{figure}[t]
\centering
\includegraphics[width=\linewidth]{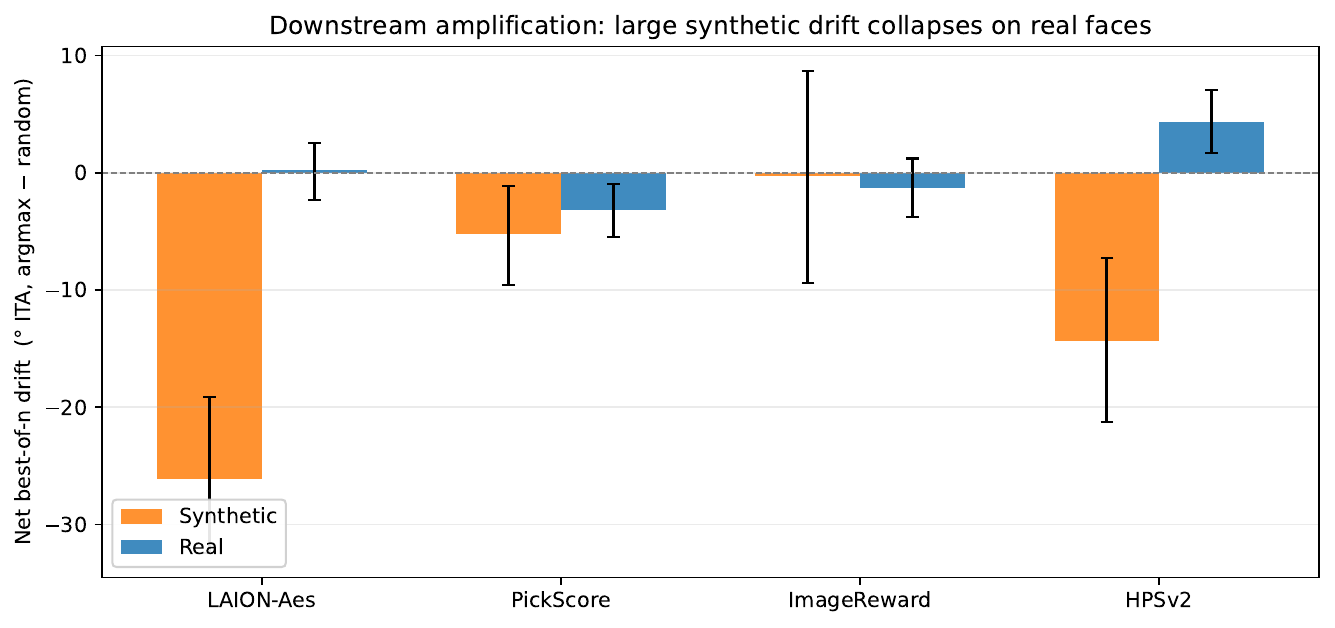}
\caption{Best-of-$n$ amplification comparison between synthetic and real arms. The
synthetic arm shows $-26.1^\circ$ ITA drift for LAION-Aes, while the real arm shows
$+0.2^\circ$ (not significant): a ${\sim}129\times$ magnitude collapse rather than
a sign flip.}
\label{fig:amplification}
\end{figure}

\paragraph{Synthetic-to-real divergence (the core of this paper).}
On real faces, LAION-Aes goes from ``preference for darker skin along the whole
dose range + $-26.1^\circ$ amplification'' to ``near-symmetric inverted U
(fidelity index $+0.074$; residual asymmetry $+0.032$, CI [$+0.005$,$+0.059$],
statistically detectable but $6.5\times$ smaller than the synthetic arm's
$-0.211$) + amplification $+0.2^\circ$, not significant.'' Same scorer, same
intervention, yet the synthetic and real readings differ in sign on the asymmetry
and by ${\sim}129\times$ in the amplification magnitude
(Figures~\ref{fig:fidelity}b and~\ref{fig:amplification}). That large directional
bias on the synthetic arm is therefore mainly a \textbf{synthetic artifact}---the
accompanying differences when SDXL renders different skin lightness, amplified by
argmax---rather than a demographic preference of the scorer toward real faces.

\paragraph{How general is the divergence? (scope of the headline.)}
We state the scope explicitly rather than let ``reversal'' stand for the whole
table, because on a strict criterion---\emph{both} arms FDR-significant
\emph{and} opposite in sign---only 4 of the 12 scorer $\times$ reading cells
qualify (2 more are ``significant $\rightarrow$ non-significant'' collapses, 2
keep the same sign and merely attenuate, and 4 are non-significant in both arms).
Table~\ref{tab:divergence} lays this out.

Three qualifications follow. (i) The \textbf{load-bearing reversal is LAION-Aes's
asymmetry}---the one cell where a large, FDR-significant synthetic effect
($-0.211$) becomes a small, oppositely-signed, FDR-significant real effect
($+0.032$). This single cell is the paper's headline, and we do not generalize
past it. (ii) HPSv2 accounts for the other three reversals, and it needs
splitting in two. On the \emph{score} readings its magnitudes are negligible
($|\mathrm{FI}| \le 0.0013$, $|\mathrm{asymmetry}| \le 0.0027$), so a sign flip
between two effects we have already declined to treat as the mechanism is not
independent corroboration. Its \emph{amplification} reversal, however, is not
negligible: $-14.3^\circ$ ITA on synthetic faces versus $+4.3^\circ$ on real
ones, both FDR-significant, i.e.\ best-of-$n$ selection moves the selected skin
tone in opposite directions on the two arms. This is genuine corroboration of the
non-transfer claim, and it shows that a score-level effect too small to matter can
still be amplified into a substantial selection-level drift---one reason we report
the two levels separately. (iii) \textbf{PickScore does not reverse on any
reading}: its fidelity index stays significantly positive and its amplification
stays significantly negative in both arms, merely shrinking ${\sim}3\times$ and
${\sim}1.6\times$. For PickScore the synthetic$\rightarrow$real story is
\emph{attenuation}, not reversal. Likewise LAION-Aes's amplification does not flip
sign---it \textbf{collapses to non-significance} ($+0.2^\circ$, CI [$-2.3$,
$+2.6$]), which is the weaker claim of ``the effect disappears,'' not ``the effect
inverts.'' The general claim licensed by this table is therefore that
\textbf{synthetic-arm effects do not transfer to real faces---sometimes
inverting, sometimes merely vanishing or attenuating---and never in a way that
would let a synthetic-only audit be trusted on direction or magnitude.}

\paragraph{Per-scorer nuance (honest presentation).}
On real faces the inverted U (fidelity preference) holds cleanly for LAION-Aes
and PickScore; ImageReward is weak (FI CI [$-0.0001$,$+0.025$] barely touches
zero). HPSv2's response to $|\Delta l|\le 15$ is \textbf{negligible in magnitude
but not zero}: its FI is $-0.0008$ with CI [$-0.0014$,$-0.0003$], i.e.\
statistically resolvable yet ${\sim}90\times$ smaller than LAION-Aes's $+0.074$
and of the \emph{opposite} sign (a slight preference for perturbed over
original). We therefore describe HPSv2 as practically insensitive at this scale
rather than unbiased, and do not count it as exhibiting the fidelity mechanism.
The residual direction of the best-of-$n$ net drift varies (PickScore
$-3.2^\circ$ toward darker, HPSv2 $+4.3^\circ$ toward lighter), all far smaller
in magnitude than the synthetic arm's $-26^\circ$. \textbf{The conclusion is not
``all scorers are unbiased,'' but ``on real faces the dominant mechanism is a
fidelity preference, and residual demographic-direction bias is weak and
scorer-dependent.''}

\subsection{Placebo Arms: Is It a Fidelity Preference, or Sensitivity to
\emph{This} Operator?}
\label{sec:placebo}

A naming objection cuts deeper than the demographic one. We call the inverted U a
``fidelity preference,'' implying \emph{any} perturbation is penalized---but the
perturbations tested so far are the attributes themselves. If a scorer were merely
sensitive to the specific pixel operator we applied, the curve would look
identical and the name would be too broad. We therefore add three \textbf{placebo
arms} on the same 210 FairFace bases, none of which changes skin tone:

\begin{itemize}
\item \textbf{\texttt{bg}}---the same CIELAB $L^*$ shift at the same doses
($|\Delta l|\le 15$), applied to the \textbf{complement} of the skin mask. This is
the decisive control: same operator, same magnitude, only the support changes.
Validation on a held-out base confirms the dual is clean (mean $|\Delta|$ in the
eroded skin core 0.0002 vs.\ 0.093--0.141 outside; measured ITA$^\circ$ moves
$-71.9^\circ \rightarrow -72.8^\circ/-71.6^\circ$).
\item \textbf{\texttt{hue}}---whole-image rotation of the CIELAB $a^*b^*$ plane by
$\pm 10/20/30^\circ$, leaving $L^*$ untouched.
\item \textbf{\texttt{blur}}---whole-image softening (Gaussian $\sigma$ up to 3 on
one side, downsample-upsample on the other), unrelated to color.
\end{itemize}

\begin{table*}[t]
\centering
\caption{Placebo arms on the same 210 FairFace bases. Same estimator, bootstrap
over base images, $B=2000$, seed 0.}
\label{tab:placebo}
\footnotesize
\begin{tabular}{lcccc}
\toprule
\textbf{Scorer} & \textbf{Skin-tone FI} & \textbf{\texttt{bg} FI [95\% CI]} &
\textbf{\texttt{hue} FI [95\% CI]} & \textbf{\texttt{blur} FI [95\% CI]} \\
\midrule
LAION-Aes & $+0.0735$ & $\mathbf{+0.0596}$ [$+0.0409$, $+0.0785$] & $+0.0319$
[$+0.0169$, $+0.0468$] & $\mathbf{+0.4359}$ [$+0.4067$, $+0.4654$] \\
PickScore & $+0.0883$ & $\mathbf{+0.1075}$ [$+0.0850$, $+0.1307$] & $+0.0951$
[$+0.0809$, $+0.1092$] & $-0.3542$ [$-0.3798$, $-0.3293$] \\
HPSv2 & $-0.0008$ & $-0.0020$ [$-0.0026$, $-0.0013$] & $-0.0039$ [$-0.0044$,
$-0.0034$] & $-0.0132$ [$-0.0140$, $-0.0124$] \\
ImageReward & $+0.0125$ & $+0.0104$ [$-0.0044$, $+0.0266$] & $-0.0404$
[$-0.0550$, $-0.0257$] & $+0.1616$ [$+0.1376$, $+0.1856$] \\
\bottomrule
\end{tabular}
\end{table*}

(\texttt{hue}/\texttt{blur} use symmetric seven-level grids matched in
\emph{number and symmetry} of doses to the skin arm; their units are not
commensurable with $\Delta l$, so we compare only sign, significance, and
relative magnitude---not slopes.)

\paragraph{The decisive comparison supports the name.}
For the two scorers that clearly exhibit the mechanism, moving the \emph{same
operator} off the skin produces a fidelity penalty \textbf{statistically
indistinguishable in magnitude} from the skin arm: LAION-Aes ratio 0.81 (CI
[0.54, 1.19]) and PickScore 1.22 (CI [0.91, 1.62])---both CIs contain 1.
PickScore's hue arm is likewise indistinguishable (1.08, CI [0.85, 1.39]). So the
penalty is not attached to skin pixels or to skin-lightness semantics; it attaches
to \emph{departure from the original image}. This is the control that separates
``fidelity preference'' from ``skin-operator sensitivity,'' and it comes out on
the side of fidelity.

\paragraph{But it also bounds the claim, in two ways we report rather than bury.}
First, the penalty is strongly \textbf{operator-dependent in magnitude}:
LAION-Aes penalizes blur ${\sim}6\times$ more than skin-tone shift (ratio 5.93,
CI [4.78, 7.81]). ``Any perturbation is penalized'' is therefore true in
\emph{sign} but not in \emph{size}---the scorers have a perturbation-sensitivity
profile, not a uniform fidelity term. Second, and more sharply, \textbf{two of the
four scorers reverse on some operator}: PickScore \emph{rewards} strong blur (FI
$-0.354$; its curve rises at $\sigma \ge 2$ in both directions) and ImageReward
\emph{rewards} hue rotation (FI $-0.040$) while penalizing blur. HPSv2 penalizes
nothing on any arm (all four FIs negative, all $|\mathrm{FI}| \le 0.013$).
Counting cells on the strict criterion (fidelity index significantly positive,
i.e.\ CI entirely above zero), only LAION-Aes shows the penalty on all four
operators; PickScore on 3/4, ImageReward on \textbf{1/4} (only blur; its skin and
\texttt{bg} arms are not CI-resolvable and its hue arm is significantly
\emph{negative}), HPSv2 on 0/4. Counting sign alone, without requiring
significance, ImageReward would be 3/4---we report the strict count, since a
fidelity claim resting on non-resolvable cells is exactly the kind of overreach
the placebo arms exist to catch.

\paragraph{Revised claim.}
The honest statement is therefore narrower than ``these scorers prefer fidelity'':
\emph{for LAION-Aes and PickScore, penalization of within-image perturbation
generalizes beyond the audited attribute and beyond the audited operator's
support, at comparable magnitude, so the inverted U in Section~\ref{sec:skin} is
not an artifact of the skin-lightness operator; but the penalty is not
operator-uniform, and for ImageReward and HPSv2 it does not generalize.} We use
``fidelity preference'' for the two scorers where the placebo arms license it, and
describe the other two as having idiosyncratic, operator-specific sensitivity
profiles. Notably, PickScore's blur reversal is a mild reward-hacking
signature---a scorer used to filter training data that assigns \emph{higher}
preference to visibly softened images---which we flag as a finding in its own
right, independent of any fairness question.

\section{Scale Calibration: Why Synthetic Audits Mislead}
\label{sec:scale}

The synthetic/real divergence in Section~\ref{sec:skin} is not a contradiction but
decomposable:

\begin{quote}
\footnotesize
cross-image observed score diff = pure attribute effect (within-image) +
composition/rendering confound (prompt or SDXL association)

\medskip
\textbf{Synthetic arm}: pure effect (inverted U, real mechanism) is polluted by
``compositional/lighting differences when SDXL renders different skin tones'' and
amplified by argmax.

\textbf{Real arm}: no prompt-associated rendering confound, the fidelity
preference (inverted U) is laid bare.
\end{quote}

Methodological implication: an audit that only does synthetic + cross-image, or
synthetic + within-image, will (i) miss the real effect at the cross-image level
because variance drowns it, and (ii) misread ``synthetic-rendering accompanying
differences'' as ``demographic-direction bias'' at the within-image level, and
exaggerate its magnitude via best-of-$n$. \textbf{To judge the direction,
magnitude, and mechanism of scorer bias, one must do within-image causal isolation
on real data.} This is a different conclusion from ``the scorer is
unbiased''---the fidelity preference is real; it just is not a one-directional
preference along the lightness axis. We stress that this concerns \emph{our}
continuous lightness axis and says nothing about categorical group rankings
measured by other means (Section~\ref{sec:related}).

\paragraph{Scope of findings.}
Our findings are limited to the intervention protocols tested (CIELAB $L^*$ shift
for skin tone; plus the non-skin $L^*$ shift, hue rotation, and blur placebo arms
of Section~\ref{sec:placebo}, and the torso warp of Section~\ref{sec:body}). We
cannot conclude that models ``only learn local statistics'' in general---the
fidelity preference could coexist with other global features (e.g., color
histograms, overall composition) that survive our interventions. What we
demonstrate is that the inverted-U response is the dominant measured effect on
these operators, that it is not specific to the skin-lightness operator
(Section~\ref{sec:placebo}), and that it is nonetheless not operator-uniform in
magnitude---not that local statistics are the only learned features.

\section{Body Axis: A Cross-Attribute Stress Test of the Fidelity Mechanism (Not
a Body-Bias Measurement)}
\label{sec:body}

\paragraph{Positioning (important).}
The purpose of adding a second attribute axis---body type---is \textbf{not} to
claim we measured ``body demographic bias,'' but to run a \textbf{cross-attribute
stress test} of the fidelity-preference mechanism from Section~\ref{sec:skin}:
with a completely different kind of pixel perturbation (geometric rather than
lightness), does ``the original state scores highest, any perturbation is
penalized'' still hold? We do not claim body bias because the body axis has an
\textbf{endogenous methodological obstacle} absent from the skin axis: skin tone
admits a single-channel, reversible, zero-artifact $L^*$ shift intervention
(\texttt{clip\_frac}=0 for $|\Delta l|\le 15$), whereas body type has \textbf{no
equivalent zero-artifact within-image intervention}---any within-image change of
body type must push pixel geometry, and pushing geometry necessarily produces
deformation artifacts. This section quantifies that obstacle with data and
tightens the claim accordingly.

SDXL generates 6 body-type gradient levels (3 scenes $\times$ 3 seeds), then each
base image is torso-geometrically warped ($\gamma$ 0.80--1.20, 7 levels
$=54\times7=378$), and the silhouette \texttt{width\_ratio} is measured.

\begin{table*}[t]
\centering
\caption{Fidelity curve, synthetic body ($n=54$ base images $\times$ 7 $\gamma$,
referenced to the $\gamma=1.0$ original; endpoints $\gamma=0.8/1.2$). Bootstrap
over base images, $B=2000$, seed 0. \checkmark~= FDR-significant in the 24-test
FI/asymmetry grid of Section~\ref{sec:rigor}; all four fidelity indices survive,
and no asymmetry does except HPSv2's, whose magnitude is negligible.}
\label{tab:body-fidelity}
\footnotesize
\begin{tabular}{lccc}
\toprule
\textbf{Scorer} & \textbf{Fidelity index [95\% CI]} & \textbf{Asymmetry [95\%
CI]} & \textbf{Inverted U} \\
\midrule
LAION-Aes & $+0.110$ [$+0.079$, $+0.144$] \checkmark & $-0.010$ [$-0.060$,
$+0.038$] & \textbf{Yes} \\
PickScore & $+0.078$ [$+0.058$, $+0.098$] \checkmark & $+0.015$ [$-0.030$,
$+0.060$] & \textbf{Yes} \\
ImageReward & $+0.040$ [$+0.024$, $+0.057$] \checkmark & $-0.010$ [$-0.046$,
$+0.024$] & \textbf{Yes} \\
HPSv2 & $+0.0019$ [$+0.0014$, $+0.0025$] \checkmark & $+0.0037$ [$+0.0027$,
$+0.0047$] \checkmark & No (magnitude negligible) \\
\bottomrule
\end{tabular}
\end{table*}

\begin{table*}[t]
\centering
\caption{Discriminative regression separating the pure width effect from the
deformation artifact: $\mathrm{score} = \alpha + \beta\cdot\mathrm{width} +
\gamma\cdot\texttt{warp\_artifact}$.}
\label{tab:body-decomp}
\footnotesize
\begin{tabular}{lcccc}
\toprule
\textbf{Scorer} & \textbf{$\beta$(width $|$ control warp)} & \textbf{Sig} &
\textbf{$\gamma$(\texttt{warp\_artifact})} & \textbf{Sig} \\
\midrule
LAION-Aes & $-0.111$ [$-0.48$, $+0.09$] & No & $\mathbf{-0.813}$ [$-1.14$,
$-0.54$] & \checkmark \\
PickScore & $-0.252$ [$-0.63$, $-0.03$] & Yes & $\mathbf{-0.680}$ [$-0.89$,
$-0.51$] & \checkmark \\
ImageReward & $-0.131$ [$-0.38$, $+0.04$] & No & $\mathbf{-0.307}$ [$-0.49$,
$-0.15$] & \checkmark \\
HPSv2 & $+0.0001$ [$-0.005$, $+0.009$] & No & $\mathbf{-0.023}$ [$-0.027$,
$-0.018$] & \checkmark \\
\bottomrule
\end{tabular}
\end{table*}

\begin{table*}[t]
\centering
\caption{From case to criterion: the three conditions of attribute
auditability.}
\label{tab:auditability}
\footnotesize
\begin{tabular}{@{}p{0.17\textwidth}p{0.29\textwidth}p{0.17\textwidth}p{0.31\textwidth}@{}}
\toprule
\textbf{Condition} & \textbf{Meaning} & \textbf{Skin lightness} & \textbf{Body
width} \\
\midrule
\textbf{C1 composition-freezable} & Can only the attribute be changed without
touching composition/pose/background & \checkmark~same image $L^*$ shift &
$\times$~changing body type pushes pixel geometry, falls back to cross-image \\
\addlinespace[2pt]
\textbf{C2 reversible low-artifact intervention} & Is the attribute-changing
operation reversible without introducing an attribute-confounded artifact &
\checkmark~\texttt{clip\_frac}=0 ($|\Delta l|\le 15$) &
$\times$~\texttt{warp\_artifact} dominates the inverted U ($\gamma$ significant
for all 4 scorers) \\
\addlinespace[2pt]
\textbf{C3 valid continuous measurement} & Is there a monotonic, calibrated
continuous objective measure of the attribute & \checkmark~ITA$^\circ$
$R^2=0.88$, monotonic rate 98\% & $\times$~\texttt{width\_ratio} 90\%
saturated, $R^2=0.02$ \\
\bottomrule
\end{tabular}
\end{table*}

\paragraph{Cross-attribute reproduction.}
LAION-Aes/PickScore/ImageReward all show a \textbf{symmetric inverted U} on the
body axis (Table~\ref{tab:body-fidelity}), and here the symmetry claim is directly
supported: all three have significantly positive fidelity index while their
asymmetry CIs straddle zero (e.g.\ LAION-Aes FI $+0.110$ [$+0.079$,$+0.144$] vs
asymmetry $-0.010$ [$-0.060$,$+0.038$]). This is a cleaner symmetry result than
the skin axis, where LAION-Aes retained a small but resolvable residual asymmetry.
HPSv2 is again the exception in kind rather than degree: both its FI ($+0.0019$)
and asymmetry ($+0.0037$) are CI-resolvable, yet its FI is 21--57$\times$ smaller
than the other three scorers' ($+0.040$ to $+0.110$), so we read it as practically
flat rather than as exhibiting the mechanism. Its asymmetry is \emph{not}
correspondingly tiny ($+0.0037$ against LAION-Aes's $-0.010$), but that comparison
is uninformative because the other scorers' asymmetries are themselves
non-significant. \textbf{The fidelity preference reproduces across attributes and
scorers, so it is not a skin-tone special case.} Together with the placebo arms of
Section~\ref{sec:placebo} (which move the \emph{same} operator off the skin region
and obtain a penalty of indistinguishable magnitude), this answers the objection
``you only measured pixel sensitivity to skin lightness'' from two independent
directions: changing the attribute, and changing the operator's support. The
qualification of Section~\ref{sec:placebo} still applies---the penalty is not
operator-uniform in magnitude, and on the body axis, as there, HPSv2 is the scorer
for which it does not hold.

\begin{figure}[t]
\centering
\includegraphics[width=\linewidth]{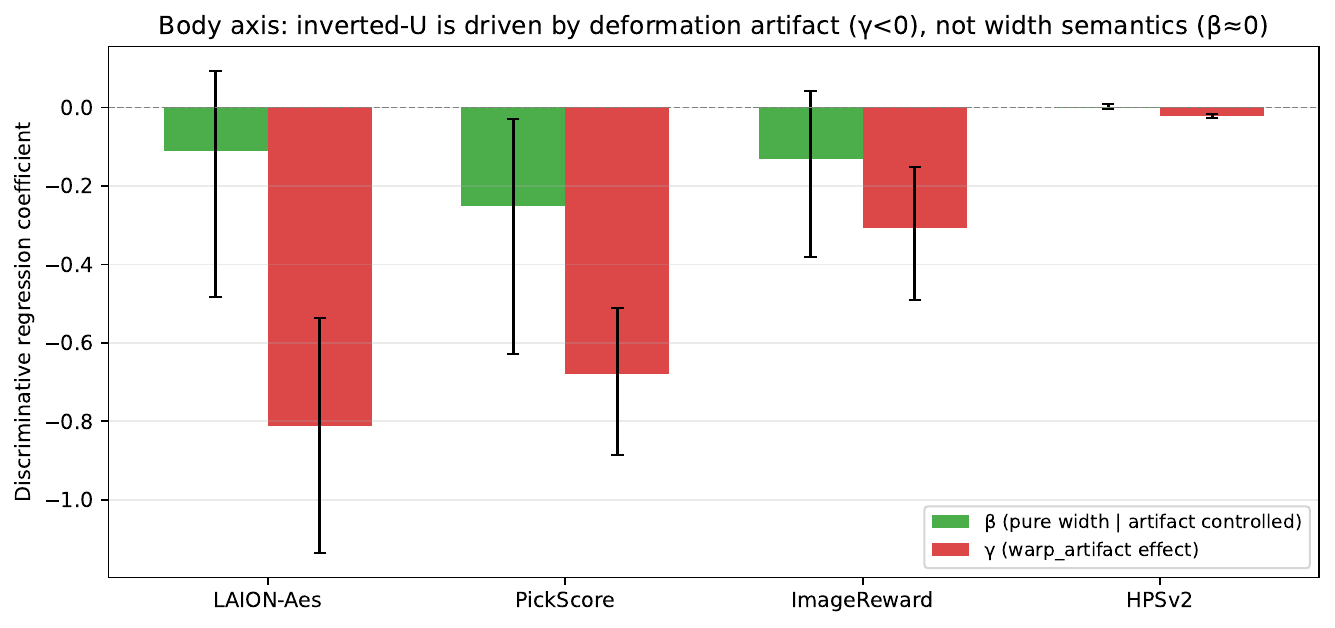}
\caption{Body axis intervention artifact decomposition. The deformation artifact
($\gamma$) dominates the inverted-U pattern while the pure width effect ($\beta$)
is weak after controlling for warp.}
\label{fig:bodydecomp}
\end{figure}

\paragraph{Necessary control for the intervention artifact (key honesty point).}
The body warp, unlike an $L^*$ shift, is not cleanly reversible---it necessarily
introduces geometric deformation. The $\gamma$ of \texttt{warp\_artifact} is
\textbf{significantly negative for all 4 scorers} (Table~\ref{tab:body-decomp}):
the deformation artifact itself depresses the score. After controlling for it, the
pure width effect $\beta$ mostly loses significance (only PickScore retains it).
Figure~\ref{fig:bodydecomp} visualizes this decomposition.

\paragraph{The inverted U is almost entirely driven by deformation magnitude
(nailing-down evidence).}
A more direct diagnostic: \texttt{warp\_artifact} is a near-perfect V-shaped
symmetry about $\gamma=1.0$ (median deformation at $\gamma$ 0.80/1.00/1.20 =
0.149 / 0 / 0.099), while after within-base demeaning:
\begin{align*}
\mathrm{corr}(\mathrm{score}, \texttt{warp\_artifact}) &= -0.396\\
\mathrm{corr}(\mathrm{score}, \texttt{width\_ratio}) &= +0.004
\end{align*}
\noindent That is, the score mainly tracks ``deformation magnitude'' and is
almost zero-correlated with true body width.
That is, the ``symmetric inverted U'' on the body axis is essentially a mirror
image of the ``V-shape of deformation magnitude,'' not any body semantics---the
within-base correlation between score and \texttt{width\_ratio} is effectively
zero. \textbf{Conclusion}: the inverted U reproduced on the body axis is
\textbf{another instance of the fidelity mechanism (penalizing any pixel
perturbation)}, and \textbf{not} a ``scorer preference for body width.'' We
accordingly tighten the claim explicitly: the body axis proves that the fidelity
preference is universal across perturbation types, but \textbf{does not claim to
have measured body demographic bias}; the latter would require a
deformation-artifact-free intervention (see Section~\ref{sec:limits} routes). This
is precisely the value of Pillar 2 (the artifact covariate)---it confirms a real
effect on the clean skin axis (\texttt{clip\_frac}=0) and actively debunks, on the
body axis, an artifact that would otherwise have been misread as ``body
preference.''

\paragraph{From case to criterion: the three conditions of attribute
auditability.}
That the skin axis can reach ``bias measurement'' while the body axis can only
reach ``mechanistic stress test'' is a difference not of data volume or effort, but
of \textbf{whether the attribute itself can be audited by pixel-level causal
isolation}. We distill this difference into a transferable criterion
(Table~\ref{tab:auditability}): all three green $\Rightarrow$ a bias measurement is
possible (skin axis, real faces $n=1470$); any one missing $\Rightarrow$ only a
mechanistic stress test (the body axis lacks all three). \textbf{This criterion is
itself a transferable contribution of the paper}: anyone wishing to audit a new
attribute (age, hairstyle, body proportion, pose) can self-check with C1/C2/C3 to
predict whether it can reach ``measurement'' or stop at ``stress test,'' rather
than discovering after the fact that artifacts polluted the conclusion. It elevates
the three-pillar benchmark of Section~\ref{sec:bench} from ``a set of tools'' to
``a map of applicability of what attributes can be audited at the pixel level.''

\paragraph{Amplification arm (body).}
random null $\approx 0$ (calibration valid); the 4 scorers' net-bias CIs all
straddle zero, not significant---consistent with the symmetric inverted U (no
directional tendency $\Rightarrow$ best-of-$n$ produces no systematic drift).

\section{Rigor Checks: Multiple Comparisons, Measurement Validity, Robustness,
Falsifiability}
\label{sec:rigor}

The credibility of audit-style empirical work rests on four things: whether
significance decisions control false positives, whether the measurement instrument
actually measures what it claims, whether conclusions are robust to randomness, and
whether conclusions are falsifiable (rather than packaging a preset stance as a
``finding''). We make all four fixed steps of the benchmark.

\paragraph{Multiple comparisons (Benjamini-Hochberg FDR).}
We run three \emph{separate} BH grids, because the readings differ in inferential
machinery and pooling them would be incoherent: (i) the 36 headline decisions below
(bootstrap CI on regression/amplification quantities plus the fidelity endpoint),
(ii) a 24-test grid over the fidelity index and asymmetry themselves (3 arms
$\times$ 4 scorers $\times$ 2 statistics; bootstrap two-sided $p$ with $+1$
smoothing, $B=2000$, so $p$ is floored at $1/(B+1)=0.0005$ and 11 of the 24 cells
report exactly that floor---they should be read as ``$p<0.001$,'' not as exact
values), and (iii) the 8 subgroup omnibus permutation tests (population
stratification, below). Grid (ii) yields \textbf{15 raw-significant, 14 significant
after FDR}; the one downgrade is the synthetic-arm ImageReward fidelity index (raw
$p=0.034 \rightarrow$ fdr $p=0.054$), which we accordingly report as
raw-significant only. Two results in grid (ii) matter for the headline: real-face
LAION-Aes asymmetry \textbf{survives FDR} (fdr $p=0.040$), which is why we say
``near-symmetric'' rather than ``symmetric''; and on the body axis all three main
scorers' asymmetry CIs straddle zero (fdr $p \ge 0.56$), so the body-axis symmetry
claim is the better-supported one. Grid (i), detailed next, produces \textbf{36
headline significance decisions} over 3 arms $\times$ 4 scorers $\times$ 3
readings. Merely checking ``whether the bootstrap 95\% CI straddles zero'' amounts
to 36 uncorrected $\alpha=0.05$ tests, with an expected ${\sim}1.8$ false
positives. For each reading we use a bootstrap two-sided $p$-value ($+1$ smoothing,
consistent with the CI convention), then apply BH-FDR ($\alpha=0.05$) jointly over
the 36 $p$-values: \textbf{22 raw-significant, 20 significant after FDR
correction.} Only 2 are downgraded from ``raw-significant''---the most typical being
the real-face LAION-Aes discriminative $\beta$ ($+0.00037$, raw-$p=0.036
\rightarrow$ not significant after FDR), which is a tiny effect to begin with, and
the correction honestly labels it ``raw-significant / not significant after FDR.''
\textbf{The main conclusions (real-face inverted-U fidelity preference, the
LAION-Aes asymmetry sign flip and amplification collapse, artifact-driven body
axis) all survive FDR correction.}

\begin{table*}[t]
\centering
\caption{Measurement-validity calibration: within-base regression of the measured
value on the known intervention dose.}
\label{tab:validity}
\footnotesize
\begin{tabular}{llcccccl}
\toprule
\textbf{Arm / axis} & \textbf{Instrument} & \textbf{Known dose} & \textbf{Slope}
& \textbf{$R^2$} & \textbf{Monotonic} & \textbf{Spearman} & \textbf{Verdict} \\
\midrule
Real face / skin & ITA$^\circ$ & $\Delta L^*$ & $+2.05$ & $\mathbf{0.790}$ &
$\mathbf{97.6\%}$ & $+0.499$ & Valid \\
Synth.\ face / skin & ITA$^\circ$ & $\Delta L^*$ & $+3.35$ & $\mathbf{0.876}$ &
$\mathbf{98.1\%}$ & $+0.734$ & Valid \\
Synthetic body & \texttt{width\_ratio} & warp $\gamma$ & $+0.066$ &
$\mathbf{0.021}$ & 66.7\% & $+0.037$ & \textbf{Invalid} \\
\bottomrule
\end{tabular}
\end{table*}

\paragraph{Measurement-validity calibration (instrument calibration).}
All causal claims rest on ``objective measurement = true attribute.'' We calibrate
the measurement instrument using the \textbf{known dose} of the within-image
intervention (Table~\ref{tab:validity}). \textbf{ITA$^\circ$ is a trustworthy
instrument on both skin arms} (real faces $R^2=0.79$ with a 97.6\% monotonic rate,
synthetic faces $R^2=0.88$ with 98.1\%)---the skin ruler that the headline
real-face conclusion depends on has a solid measurement basis.
\textbf{\texttt{width\_ratio} barely responds under the warp intervention}
($R^2=0.02$, Spearman $\approx 0$)---this independently confirms
Section~\ref{sec:body}'s judgment: the body axis cannot make quantitative causal
readings via \texttt{width\_ratio}. Measurement-validity calibration is therefore
not just a formality; it actively exposes an instrument defect on one axis.

\begin{table*}[t]
\centering
\caption{Robustness over 5 random seeds (real-face skin axis). Same-sign rate is
100\% on every reading.}
\label{tab:robustness}
\footnotesize
\begin{tabular}{lccc}
\toprule
\textbf{Scorer} & \textbf{Fidelity index (mean$\pm$std, sig rate)} &
\textbf{Discrim.\ $\beta$ (sig rate)} & \textbf{Net drift (mean$\pm$std, sig
rate, same-sign)} \\
\midrule
LAION-Aes & $+0.0735\pm0.0000$ (100\%) & $+0.0004$ (100\%) &
$+0.19^\circ\pm0.09$ (\textbf{0\%}, 100\%) \\
PickScore & $+0.0883\pm0.0000$ (100\%) & $-0.0003$ (100\%) &
$-3.18^\circ\pm0.12$ (100\%, 100\%) \\
ImageReward & $+0.0125\pm0.0000$ (100\%) & $+0.0003$ (0\%) &
$-1.22^\circ\pm0.11$ (0\%, 100\%) \\
HPSv2 & $-0.0008\pm0.0000$ (100\%) & $+0.0000$ (100\%) &
$+4.33^\circ\pm0.10$ (100\%, 100\%) \\
\bottomrule
\end{tabular}
\end{table*}

\paragraph{Robustness (multiple random seeds).}
The three headline readings differ in seed sensitivity: the point estimates of the
fidelity index and discriminative $\beta$ are deterministic quantities
(within-base paired differences / OLS), where seeds only affect the bootstrap
significance decision; the point estimate of the best-of-$n$ net drift itself
depends on random subsampling, which most tests robustness.
Table~\ref{tab:robustness} repeats the analysis over 5 seeds. \textbf{All readings
have a 100\% same-sign rate}---the direction is fully stable to random seeds. The
small-magnitude net drifts (LAION-Aes $+0.19^\circ$, ImageReward $-1.22^\circ$)
have a 0\% significance rate, i.e., ``direction stable but underpowered,'' which we
accordingly report as an \textbf{honest null rather than neutrality}. This forms a
sharp contrast with the synthetic-arm LAION-Aes $-26.1^\circ\pm0.24$ (100\%
significance rate), further confirming ``the large synthetic drift is an artifact,
the real drift is weak.''

\paragraph{Skin-segmentation threshold sensitivity.}
A natural objection is: ITA$^\circ$ depends on YCbCr skin-segmentation thresholds,
so might the conclusion be merely an artifact of one threshold set? The key is that
\textbf{the headline conclusion is inherently immune to thresholds}---the
inverted-U fidelity curve uses the within-base paired difference of score on the
\textbf{known $\Delta l$} (the offset applied to CIELAB $L^*$), and $\Delta l$ is an
intervention parameter that does not pass through skin segmentation; ITA$^\circ$
involves thresholds only when it is the independent variable in the discriminative
regression $\beta(\mathrm{ITA}|\mathrm{clip})$. We stress-test this one affected
reading: re-segment, recompute ITA, and rerun the within-base discriminative
regression on all 210 real-face bases (1470 images) with tightened / default /
loosened YCbCr threshold sets. PickScore (the one FDR-significant real-arm $\beta$)
stays negative and within 10\% of its own magnitude across all three sets
(tightened $-0.00036$, default $-0.00035$, loosened $-0.00033$); it is
CI-resolvable at the tightened and default sets and marginally straddles zero at
the loosened one (CI upper bound $+0.00002$). LAION-Aes's near-zero $\beta$
likewise keeps one sign across all three (tightened $+0.00041$, default
$+0.00037$, loosened $+0.00024$), CI-resolvable only at the default set and not
significant after full FDR---consistent with it being weak to begin with. So
\textbf{the sign of the directed effect does not flip with thresholds, and its
magnitude moves by less than the width of its own CI}, though the marginal cells
cross the significance line as the mask loosens: threshold choice moves the
\emph{power}, not the \emph{direction}. The default-threshold $\beta$s recomputed
here reproduce the Section~\ref{sec:skin} table values exactly, confirming that the
two pipelines share one ITA implementation. One caveat we hit ourselves: this check
is only meaningful at the full base count. On 30- and 60-base subsets no cell
reaches significance and the estimated sign is itself unstable, so the ``subsets
suffice for sign stability'' shortcut in our own probe documentation was wrong and
we removed it.

\paragraph{Falsifiability: this is not drawing our own target.}
An honest validity threat is: could ``fidelity preference'' be a conclusion we
thought up in advance and then back-derived using cherry-picked
interventions/measurements (circular reasoning / straw man)? We respond not with
argument but by pointing out that this design's \textbf{result space is open, and
could have been falsified in several places but was not}---if the conclusion were
constructed, the following observations would not appear:

\begin{itemize}
\item \textbf{The two arms could have agreed but conflicted.} If we only wanted to
``prove a fidelity preference,'' the easiest path would be to have the synthetic and
real arms give a consistent inverted U. The reality is the opposite: on the
synthetic arm LAION-Aes tilts \textbf{toward darker across the dose range}
(asymmetry $-0.211$; fidelity index $-0.005$, itself not significant, i.e.\ no
inverted U at all; amplification $-26.1^\circ$), and on the real arm it becomes a
\textbf{near-symmetric inverted U} (fidelity index $+0.074$; residual asymmetry
$+0.032$, small but CI-resolvable and oppositely signed; amplification
$+0.2^\circ$, not significant). A preset target would not produce two arms whose
asymmetries are \textbf{opposite in sign} and whose amplifications differ by
${\sim}129\times$ in magnitude---it was the data that forced us to distinguish
``synthetic artifact'' from ``real mechanism,'' not our choice of that distinction.
\item \textbf{The validity calibration could have all passed but actively vetoed
one axis.} Measurement-validity calibration passes for skin ITA$^\circ$
($R^2=0.79/0.88$) but \textbf{rules \texttt{width\_ratio} invalid} for body
($R^2=0.02$). If the criterion existed to make the conclusion look good, it would
not retain a step that vetoes its own main measurement. We accordingly
\textbf{actively tighten} the body-axis claim (from ``body bias'' down to
``mechanistic stress test'')---a narrowing, not a broadening, of the conclusion.
\item \textbf{Significance could have reported only raw $p$ but ran an FDR that
hurts itself.} BH-FDR downgrades the real-face LAION-Aes discriminative $\beta$
from ``raw-significant'' to not significant. A self-target analysis would not
introduce a correction that \textbf{downgrades its own headline candidate}.
\item \textbf{The null arm could have been read as ``unbiased'' but was reported as
``underpowered.''} Most of the best-of-$n$ real-arm net drifts have CIs straddling
zero. We did not package this as ``the scorer is unbiased toward real
populations'' (which would be prettier), but combined it with robustness
(same-sign rate 100\%, sig rate 0\%) to honestly report ``direction stable but
underpowered.''
\end{itemize}

In other words, \textbf{the core empirical findings come from an open result space
that permits falsification}: the conditions gave conflicting results (synthetic
monotonic vs real inverted U), the calibration passed one axis and vetoed another,
and the correction hurt its own candidate conclusion. A self-drawn target cannot
satisfy all these constraints simultaneously. We fix this section into the
benchmark precisely so that ``is the conclusion constructed'' \textbf{can be
re-tested for every new attribute/scorer}, rather than relying on the authors to
vouch for themselves.

\paragraph{Population stratification: is the fidelity preference symmetric across
populations?}
The fidelity preference ``looks population-neutral,'' but the harm hides in two
population asymmetries, visible only when the fidelity curve is stratified by
FairFace's built-in population labels: (a) population differences in
\textbf{penalty strength} (if the fidelity index varies by population, ``deviation
from the original state'' is penalized more heavily for some populations), and (b)
population differences in \textbf{residual directional bias} (if the asymmetry is
significantly positive/negative for some population, that population has a
lightness-direction bias averaged out by the overall symmetric inverted U). The
within-base paired difference naturally cancels the base-image intercept, so the
population contrast is not polluted by differences in photo quality across groups.
\textbf{The key methodological trap}: 7 populations yield 21 pairwise contrasts, and
reporting only ``the most different pair'' (disparity) amounts to cherry-picking
extremes and systematically overstates significance---precisely what the
falsifiability spirit of this section guards against. We therefore take a
\textbf{selection-free permutation test (omnibus)} as the standard: shuffle
population labels and recompute the between-group max$-$min spread, checking whether
the observed value exceeds the random null; then apply BH-FDR jointly over the 4
scorers $\times$ 2 metrics = 8 omnibus $p$-values.

The result sharply demonstrates the gap between ``cherry-picking extremes'' and
``selection-free'': LAION-Aes's largest directional-bias contrast disparity looks
significant (East Asian$-$Indian, raw-$p=0.014$), but its omnibus $p=0.422$---purely
a cherry-picking artifact. \textbf{Of the 8 omnibus tests, only 2 survive FDR, both
from HPSv2} (penalty-strength heterogeneity fdr-$p=0.024$, directional-bias
heterogeneity fdr-$p=0.024$); PickScore's penalty-strength heterogeneity drops to
marginal (omnibus raw $0.021 \rightarrow$ fdr $0.057$, does not pass),
ImageReward's directional-bias heterogeneity raw $0.050 \rightarrow$ fdr $0.100$,
and LAION-Aes's two items are both far from significant (Table~\ref{tab:subgroup}).

\begin{table*}[t]
\centering
\caption{Population stratification: selection-free omnibus permutation tests,
BH-FDR over 4 scorers $\times$ 2 metrics. Only HPSv2 survives.}
\label{tab:subgroup}
\footnotesize
\begin{tabular}{lccl}
\toprule
\textbf{Scorer} & \textbf{Penalty-strength omnibus (fdr)} &
\textbf{Directional-bias omnibus (fdr)} & \textbf{Conclusion} \\
\midrule
LAION-Aes & 0.422 (0.483) & 0.155 (0.249) & population differences not robust \\
PickScore & \textbf{0.021} (0.057) & 0.347 (0.462) & strength marginal (fails),
no direction \\
ImageReward & 0.613 (0.613) & 0.050 (0.100) & direction marginal, no strength \\
HPSv2 & \textbf{0.006 (0.024 \checkmark)} & \textbf{0.004 (0.024 \checkmark)} &
\textbf{population differences robust} \\
\bottomrule
\end{tabular}
\end{table*}

\begin{figure}[t]
\centering
\includegraphics[width=\linewidth]{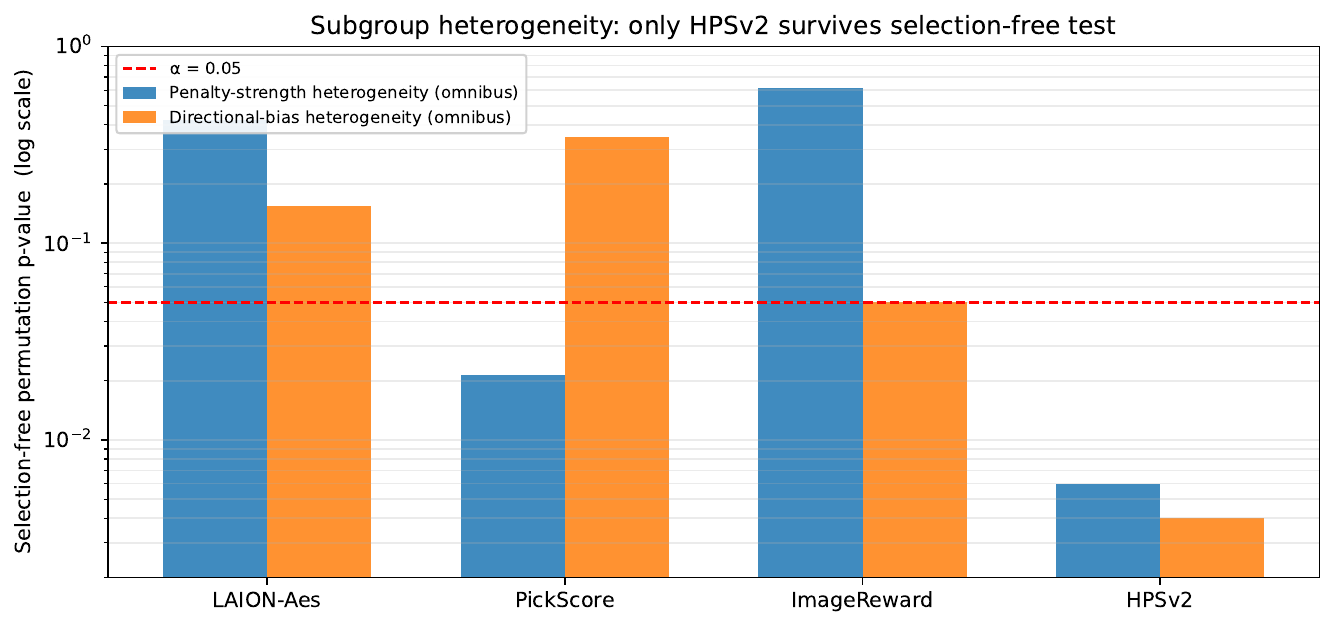}
\caption{Population stratification omnibus test results. Comparing disparity
(largest pairwise contrast) against selection-free omnibus $p$-values demonstrates
that most apparent population differences are cherry-picking artifacts. Only HPSv2
survives FDR correction.}
\label{fig:subgroup}
\end{figure}

\paragraph{This result is both a finding and a methodological demonstration.}
As a finding: the harm of ``fidelity penalty systematically varying by population''
on real faces is, on most scorers, \textbf{not robust} after a selection-free test +
FDR---we do not inflate it into ``the scorer treats populations differently''; the
only robust exception is HPSv2, which, though overall insensitive to pixel-level
lightness (Section~\ref{sec:skin}), does have its residual tiny directional bias
vary systematically by population (e.g., asymmetry significantly positive for the
white/Middle-Eastern levels, negative for the Black level), a signal worth tracking
separately. As a demonstration: the gap between LAION-Aes's disparity (0.014
``significant'') and omnibus (0.422 not significant) is a live sample of this
section's falsifiability pillar catching a false positive---had we followed the
common practice of ``report the largest contrast,'' we would have wrongly declared a
population bias that does not exist. Figure~\ref{fig:subgroup} visualizes the
omnibus test results.

\section{Deliverable: The Reproducible Audit Benchmark \texttt{audit\_bench}}
\label{sec:deliverable}

\begin{table}[t]
\centering
\caption{Structure of the released benchmark.}
\label{tab:repo}
\footnotesize
\begin{tabular}{@{}p{0.27\columnwidth}p{0.67\columnwidth}@{}}
\toprule
\textbf{Path} & \textbf{Role} \\
\midrule
\texttt{config.py} & Single source of truth: 4 scorers $\times$ 2 architecture
families, 2 axes, 3 arms, paths/environment \\
\addlinespace[2pt]
\texttt{orchestr\-ator.py} & One-click pipeline: scoring $\rightarrow$ three
probes $\rightarrow$ report (\texttt{-\--analyze-only} for second-scale
reproduction) \\
\addlinespace[2pt]
\texttt{report.py} & Orchestration of three probes $\rightarrow$
\texttt{audit\_report.json} \\
\addlinespace[2pt]
\texttt{PROTOCOL.md} & Methodology: within-image intervention + artifact
calibration + best-of-$n$ null + fidelity curve \\
\bottomrule
\end{tabular}
\end{table}

\begin{itemize}
\item \textbf{Within-image intervention protocol}: pixel-level CIELAB lightness
intervention (skin) / torso geometric warp (body), feather/window-function
blending, composition frozen.
\item \textbf{Intervention-artifact covariate}: \texttt{clip\_frac}
(over-clipping) / \texttt{warp\_artifact} (deformation), used as covariates in the
discriminative regression to separate attribute semantics from intervention
side-effects.
\item \textbf{Fidelity curve}: within-base paired difference referenced to the
original $\rightarrow$ fidelity index + asymmetry, detecting the inverted U and
distinguishing ``fidelity preference'' from ``demographic-direction preference.''
\item \textbf{best-of-$n$ null calibration}: random baseline + paired bootstrap,
net bias = argmax $-$ random.
\item \textbf{Dual-arm cross-validation}: synthetic + real side by side, exposing
the non-transfer of synthetic effects (sign flip for LAION-Aes's asymmetry,
collapse for its amplification).
\item \textbf{Placebo arms} (\fpath{interventions/\bk placebo.py},
\fpath{probes/\bk placebo\_\bk contrast.py}): the same operator applied off the
attribute's support (\texttt{bg}), plus two unrelated operators (\texttt{hue},
\texttt{blur}), with a paired ratio bootstrap against the attribute arm. This is
what licenses---or refuses---the word ``fidelity'' for a given scorer, and it
generalizes to any new attribute axis: whoever adds one should add its
complement-support placebo before naming the mechanism.
\item Unified scorer interface \fpath{score(images)\bk ->list[float]}; adding a new
scorer/attribute axis only requires editing config. One-click:
\fpath{python -m \bk audit\_bench.\bk orchestrator \bk --analyze-only}.
\end{itemize}

\section{Limitations and Future Work}
\label{sec:limits}

\begin{enumerate}
\item \textbf{Fairness reading of the fidelity preference}. The fidelity preference
(inverted U) is not a traditional one-directional demographic bias; its harm
mechanism is ``reward the original state closest to the training-distribution mode,
penalize the long tail,'' and it constitutes demographic harm only when combined
with group-differentiated penalties on the original state.
Section~\ref{sec:rigor}'s population stratification has already taken this
population-level quantification step using the existing 1470 real faces: the
conclusion is that \textbf{population heterogeneity is, on most scorers, not robust
after a selection-free test + FDR, with only HPSv2 surviving}. This converges the
harm from ``theoretically possible'' to ``measured, and not a robust
population-level harm on most scorers,'' but the detection power of 30
bases/population is limited, and population-level harm quantification at larger
scale and with more attribute crossings (skin tone $\times$ body type) is left for
future work.

\item \textbf{The endogenous methodological obstacle of the body axis (the most
important boundary of this paper, not a data-volume issue)}. The skin-axis $L^*$ is
a single-channel, reversible, zero-artifact intervention (\texttt{clip\_frac}=0 for
$|\Delta l|\le 15$); the body axis has \textbf{no equivalent intervention}---any
within-image change of body type must push pixel geometry and necessarily produce
\texttt{warp\_artifact}. Section~\ref{sec:body} has quantified this: the
within-group correlation of score with deformation magnitude is $-0.40$, with true
width $+0.004$, and the inverted U is almost entirely a mirror of deformation.
\textbf{Key clarification: this is not a problem that ``lacking a real arm'' can
solve.} Adding real-model images removes the SDXL rendering confound, but as long as
a warp is still used to create within-image gradients, the deformation artifact
remains. We therefore claim only ``the fidelity mechanism reproduces across
perturbation types'' for the body axis, not ``body demographic bias.'' Two
root-cause routes:

\begin{itemize}
\item \textbf{Route A (natural body-type cross-image arm)}: recruit real models
across the body-type spectrum, shooting with fixed background/pose/clothing/camera
---body type is the models' innate attribute rather than warped, so the
\texttt{warp\_artifact} confound vanishes entirely and \texttt{width\_ratio} is the
true body type. The cost is falling back to a paired cross-image design (different
people), requiring the same clothing + pose standardization + covariates to suppress
compositional confounds. This arm is led by domain collaborators for data collection
(spec in the appendix).
\item \textbf{Route B (deformation-free within-image intervention)}: use a
parametric 3D human body (SMPL-like) or diffusion semantic editing to re-render the
same person with different body types, changing body type without leaving stretch
artifacts. On assessment we consider \textbf{this route still cannot yield a clean
measurement under the ``no live studio shoot'' constraint}, and its cost is more
hidden than Route A's: (i) a pure SMPL mesh is a textureless gray model, and
CLIP-family aesthetic scorers are severely OOD on gray models with meaningless
scores; adding realistic texture/mapping reintroduces new appearance confounds; (ii)
diffusion semantic editing (SDXL inpaint/img2img), when changing body type, has the
model's prior about ``fat/thin'' incidentally alter the face, pose, clothing, and
background---without skeleton locking (ControlNet/keypoints), it swaps ``warp
deformation artifact'' for a harder-to-quantify ``diffusion semantic drift,''
violating C1 (composition freezing) and C2 (low artifact). Route B is thus not
``purification'' but swapping one confound for another; only with skeleton locking +
texture-consistency constraints could it approach usability, with a substantially
heavier tech stack.
\end{itemize}

\textbf{Exploratory pre-validation (SSP-3D).} We used
SSP-3D~\cite{sengupta2020synthetic} (311 real full-body athlete photos + SMPL shape
parameters $\beta$, 62 unique body-type individuals) to pilot a prototype of Route
A---real photos, natural body types, \textbf{no warp artifact}. Three findings
calibrated the design of the formal collection: (i) \textbf{\texttt{width\_ratio}
fails completely on uncontrolled real photos}: 90\% of samples saturate to 1.0,
correlate only $-0.15$ with SMPL $\beta_1$, confirming that under ``full-frame/free
composition'' one must switch to a controlled camera setup or use direct
anthropometry; the main dose was therefore switched to SMPL $\beta_1$; (ii)
\textbf{one must aggregate by individual to prevent pseudo-replication}: video frame
sampling makes 311 frames contain only 62 individuals, and frame-level regression
inflates PickScore/ImageReward effects into ``false significance,'' which recede to
non-significant after individual-level aggregation ($n=62$)---this is the source of
the hard constraint ``each body type is an independent individual'' for Route A
collection; (iii) \textbf{the directional preference on real body types disappears
after controlling for observable confounds} (covariate accounting ladder,
individual-level $n=62$): among the raw $\beta_1$ slopes, LAION-Aes ($-0.067$) and
HPSv2 ($-0.0012$) are significant, PickScore/ImageReward are not; but as the
regression progressively controls for height (SMPL $\beta_0$), sex, and sport
category, both originally significant scorers' $\beta_1$ CIs turn to straddle zero
(LAION-Aes $-0.067 \rightarrow -0.024$, HPSv2 $-0.0012 \rightarrow -0.0009$),
mechanistically attributable to the collinearity between body type and sport
($\mathrm{corr}(\beta_1, \text{strength sport})=-0.50$; weightlifters/powerlifters
are innately more muscular), i.e., about half of the directional signal in the raw
correlation is propped up by the sport confound. The directional preference
disappearing after controlling for confounds contrasts with the synthetic warp
arm's ``strong symmetric inverted U'' and is consistent with the skin-axis
conclusion ``demographic-direction bias on real data is weak.'' This arm is a
heavily-confounded exploratory pilot (in-the-wild composition/clothing/sport all
differ), reading direction only and making no causal claim, but it turns Route A
from a ``proposal'' into an ``executable plan with known design constraints.''

\item \textbf{Body-measurement saturation}. The silhouette \texttt{width\_ratio}
tops out under full-frame composition ($\sim$23\% of samples $\approx$1.0),
weakening the detection power of the body discriminative $\beta$ (dose monotonicity
is still preserved); switching to keypoints/segmentation would improve this.

\item \textbf{Semantic narrowing of the pixel intervention}. The $L^*$ shift only
changes skin lightness, not other ethnicity-related morphology; it isolates ``pixel
sensitivity to skin lightness,'' which is not the full bias toward ``ethnicity.''
This is a deliberate narrowing in exchange for clean causal identification---the
title and conclusions are strictly bounded within it.

\item \textbf{Coverage of real-data axes}. The real arm currently covers only skin
tone (FairFace, 1470 images); the body axis currently has only a synthetic arm,
limited by the endogenous obstacle above. The Route A real-body arm is the most
important follow-up and the only clean path to upgrade the body axis from
``mechanistic stress test'' to ``body-bias measurement.''

\item \textbf{Synthetic ecological validity is already reinforced by the real
arm}. The core conclusions are anchored on 1470 real faces ($4\times$ the synthetic
arm); the synthetic arm mainly serves to reveal the methodological point ``naive
synthetic audits mislead.''
\end{enumerate}

\paragraph{Methodological reusability.}
This benchmark of within-image causal isolation + intervention-artifact calibration
+ best-of-$n$ null calibration + fidelity curve can transfer to any audit scenario
that ``uses a learned scalar scorer to filter/rerank generated content'' (video
aesthetics, speech naturalness, text reward models), as a general template for
distinguishing ``content confound / intervention artifact / fidelity preference /
true demographic bias.''

\section{Ethics Statement and Broader Impact}
\label{sec:ethics}

\paragraph{Research positioning and potential benefits.}
This paper audits the aesthetic/preference scorers widely deployed in T2I
training-data filtering, best-of-$n$ sampling, and RLHF alignment, aiming to turn
``whether a scorer encodes demographic attributes as quality'' from a qualitative
concern into a reproducible quantitative diagnosis. Its positive impact is to give
dataset builders and model developers an actionable validity-check tool and to
correct a real methodological misconception---naive synthetic audits misjudge both
the direction and magnitude of bias (Section~\ref{sec:scale}).

\paragraph{Data sources and licensing.}
This paper uses only public datasets and collected no new human-subject data. (a)
\textbf{FairFace} (real-face skin arm, 1470 images) is used for bias-measurement
research under its release license; we use only its images and built-in race/gender
labels, restricted to population-stratified validity analysis
(Section~\ref{sec:rigor}), and train no face-recognition or attribute-prediction
model. (b) \textbf{SSP-3D} (exploratory body arm, 311 frames / 62 individuals) is
used under its research-use terms, reading only its RGB images and SMPL shape
parameters, with no identity association. (c) Synthetic-arm images are generated by
\textbf{SDXL} and contain no real personal identities. All data are stored locally,
with no images or personal data transmitted to any third-party endpoint.

\paragraph{Stance on the use of demographic labels.}
We characterize skin tone with the dermatological continuous scale ITA$^\circ$ and
use race labels only as a stratification variable, in order to \textbf{measure and
constrain} the scorer's differential treatment of populations, not to build or
reinforce any population-classification system. ITA$^\circ$ isolates only ``pixel
sensitivity to skin lightness'' and is not a full characterization of ``ethnicity''
(Section~\ref{sec:limits}, limitation 4); the race labels follow FairFace's original
categories, and we do not claim them as an essentialist population division. The
body axis uses de-stigmatized neutral descriptions and objective shape parameters
throughout, making no aesthetic value judgment.

\paragraph{Harm discussion.}
(a) \textbf{Dual-use}: the ``fidelity preference'' mechanism revealed here could in
principle be reverse-engineered to amplify rather than mitigate bias; by fully
open-sourcing the audit protocol and stressing the necessity of within-image
isolation on real data, we make defensive use (validity checking) easier than
offensive use. (b) \textbf{No exaggeration}: we deliberately avoid packaging weak
evidence as strong conclusions---residual demographic-direction bias on real faces
is weak and scorer-dependent (Section~\ref{sec:skin}), population heterogeneity is
mostly not robust after a selection-free test + FDR (Section~\ref{sec:rigor}), and
the body axis makes no bias-measurement claim (Section~\ref{sec:body}). Overclaiming
that a scorer ``severely discriminates'' is itself a harm. (c) \textbf{Scope of
applicability}: conclusions are strictly bounded to the 4 scorers, 2 attribute axes,
and neutral-prompt condition tested; they should not be extrapolated to a wholesale
judgment of unaudited scorers or real deployment pipelines.

\paragraph{Reproducibility and compliance.}
Full code, config, and structured results are released with the paper (see
reproduction checklist). Because no new human-subject collection is involved, this
study does not trigger institutional ethics review; the future real-body arm
collection (Route A) involves human subjects and will complete informed consent,
compensation, de-identification, and institutional ethics review before collection
(spec in the appendix \fpath{REAL\_\bk BODY\_\bk DATA\_\bk SPEC.md}).

\section*{Generative AI Usage Statement}

Generative AI tools were used as a writing and coding aid in preparing this work:
for drafting and copy-editing prose, for translating between the Chinese and English
versions of the manuscript, and for implementing and refactoring analysis code. All
experimental design decisions, all statistical procedures, and all substantive claims
are the authors' own. Every number reported in this paper is machine-verified against
the released structured results by \texttt{verify\_ci.py}, which re-derives each
value from \texttt{results/} and exits non-zero on any mismatch; no reported quantity
originates from a generative model. The authors have read and verified the final
manuscript in full and take responsibility for its content.

\bibliographystyle{ACM-Reference-Format}
\bibliography{refs}

@inproceedings{schuhmann2022laion,
  author    = {Schuhmann, Christoph and Beaumont, Romain and Vencu, Richard and
               Gordon, Cade and Wightman, Ross and Cherti, Mehdi and Coombes, Theo
               and Katta, Aarush and Mullis, Clayton and Wortsman, Mitchell and
               Schramowski, Patrick and Kundurthy, Srivatsa and Crowson, Katherine
               and Schmidt, Ludwig and Kaczmarczyk, Robert and Jitsev, Jenia},
  title     = {{LAION-5B}: An Open Large-Scale Dataset for Training Next
               Generation Image-Text Models},
  booktitle = {Advances in Neural Information Processing Systems 35 (NeurIPS 2022),
               Datasets and Benchmarks Track},
  pages     = {25278--25294},
  year      = {2022},
  note      = {arXiv:2210.08402. The LAION-Aesthetics predictor was released
               alongside this dataset}
}

@inproceedings{kirstain2023pick,
  author    = {Kirstain, Yuval and Polyak, Adam and Singer, Uriel and Matiana,
               Shahbuland and Penna, Joe and Levy, Omer},
  title     = {Pick-a-Pic: An Open Dataset of User Preferences for Text-to-Image
               Generation},
  booktitle = {Advances in Neural Information Processing Systems 36 (NeurIPS 2023)},
  pages     = {36652--36663},
  year      = {2023},
  note      = {arXiv:2305.01569 (PickScore)}
}

@inproceedings{xu2023imagereward,
  author    = {Xu, Jiazheng and Liu, Xiao and Wu, Yuchen and Tong, Yuxuan and
               Li, Qinkai and Ding, Ming and Tang, Jie and Dong, Yuxiao},
  title     = {{ImageReward}: Learning and Evaluating Human Preferences for
               Text-to-Image Generation},
  booktitle = {Advances in Neural Information Processing Systems 36 (NeurIPS 2023)},
  pages     = {15903--15935},
  year      = {2023},
  note      = {arXiv:2304.05977}
}

@article{wu2023hps,
  author  = {Wu, Xiaoshi and Hao, Yiming and Sun, Keqiang and Chen, Yixiong and
             Zhu, Feng and Zhao, Rui and Li, Hongsheng},
  title   = {Human Preference Score v2: A Solid Benchmark for Evaluating Human
             Preferences of Text-to-Image Synthesis},
  journal = {arXiv preprint arXiv:2306.09341},
  year    = {2023},
  note    = {Checked 2026-08-13: still preprint-only, no peer-reviewed venue}
}

@inproceedings{kusner2017counterfactual,
  author    = {Kusner, Matt J. and Loftus, Joshua and Russell, Chris and
               Silva, Ricardo},
  title     = {Counterfactual Fairness},
  booktitle = {Advances in Neural Information Processing Systems 30 (NeurIPS 2017)},
  year      = {2017}
}

@article{chardon1991skin,
  author  = {Chardon, A. and Cretois, I. and Hourseau, C.},
  title   = {Skin Colour Typology and Suntanning Pathways},
  journal = {International Journal of Cosmetic Science},
  volume  = {13},
  number  = {4},
  pages   = {191--208},
  year    = {1991},
  doi     = {10.1111/j.1467-2494.1991.tb00561.x},
  note    = {Source of the ITA$^\circ$ individual typology angle scale}
}

@inproceedings{bianchi2023easily,
  author    = {Bianchi, Federico and Kalluri, Pratyusha and Durmus, Esin and
               Ladhak, Faisal and Cheng, Myra and Nozza, Debora and
               Hashimoto, Tatsunori and Jurafsky, Dan and Zou, James and
               Caliskan, Aylin},
  title     = {Easily Accessible Text-to-Image Generation Amplifies Demographic
               Stereotypes at Large Scale},
  booktitle = {Proceedings of the 2023 ACM Conference on Fairness, Accountability,
               and Transparency (FAccT '23)},
  pages     = {1493--1504},
  year      = {2023},
  doi       = {10.1145/3593013.3594095}
}

@inproceedings{luccioni2023stable,
  author    = {Luccioni, Alexandra Sasha and Akiki, Christopher and
               Mitchell, Margaret and Jernite, Yacine},
  title     = {Stable Bias: Analyzing Societal Representations in Diffusion Models},
  booktitle = {Advances in Neural Information Processing Systems 36 (NeurIPS 2023),
               Datasets and Benchmarks Track},
  year      = {2023},
  note      = {Spotlight. arXiv:2303.11408}
}

@inproceedings{podell2024sdxl,
  author    = {Podell, Dustin and English, Zion and Lacey, Kyle and
               Blattmann, Andreas and Dockhorn, Tim and M{\"u}ller, Jonas and
               Penna, Joe and Rombach, Robin},
  title     = {{SDXL}: Improving Latent Diffusion Models for High-Resolution Image
               Synthesis},
  booktitle = {International Conference on Learning Representations (ICLR 2024)},
  year      = {2024},
  note      = {Spotlight. arXiv:2307.01952}
}

@inproceedings{karkkainen2021fairface,
  author    = {K{\"a}rkk{\"a}inen, Kimmo and Joo, Jungseock},
  title     = {{FairFace}: Face Attribute Dataset for Balanced Race, Gender, and
               Age for Bias Measurement and Mitigation},
  booktitle = {2021 IEEE Winter Conference on Applications of Computer Vision (WACV)},
  pages     = {1547--1557},
  year      = {2021},
  doi       = {10.1109/WACV48630.2021.00159}
}

@inproceedings{sengupta2020synthetic,
  author    = {Sengupta, Akash and Budvytis, Ignas and Cipolla, Roberto},
  title     = {Synthetic Training for Accurate 3D Human Pose and Shape Estimation
               in the Wild},
  booktitle = {British Machine Vision Conference (BMVC)},
  year      = {2020},
  doi       = {10.5244/C.34.20},
  note      = {arXiv:2009.10013 (SSP-3D dataset)}
}

@inproceedings{radford2021clip,
  author    = {Radford, Alec and Kim, Jong Wook and Hallacy, Chris and
               Ramesh, Aditya and Goh, Gabriel and Agarwal, Sandhini and
               Sastry, Girish and Askell, Amanda and Mishkin, Pamela and
               Clark, Jack and Krueger, Gretchen and Sutskever, Ilya},
  title     = {Learning Transferable Visual Models from Natural Language
               Supervision},
  booktitle = {Proceedings of the 38th International Conference on Machine
               Learning (ICML 2021)},
  series    = {PMLR},
  volume    = {139},
  pages     = {8748--8763},
  year      = {2021}
}

@inproceedings{abdelmagid2026bias,
  author    = {Abdel Magid, Salma and Guo, Guanghao and Tureci, Ege and
               Dharmasiri, Aniruddha and Ramaswamy, Vikram V. and
               Pfister, Hanspeter and Russakovsky, Olga},
  title     = {Bias at the End of the Score},
  booktitle = {IEEE/CVF Conference on Computer Vision and Pattern Recognition
               (CVPR)},
  year      = {2026},
  note      = {arXiv:2604.13305}
}

@inproceedings{taylor2026gaze,
  author    = {Taylor, Jared and Agnew, William and Sap, Maarten and
               Fox, Sarah E. and Zhu, Haiyi},
  title     = {The Algorithmic Gaze of Image Quality Assessment: An Audit and
               Trace Ethnography of the {LAION}-Aesthetics Predictor},
  booktitle = {Proceedings of the 2026 ACM Conference on Fairness, Accountability,
               and Transparency (FAccT '26)},
  year      = {2026},
  note      = {arXiv:2601.09896}
}

@article{guo2025universal,
  author  = {Guo, Wei Ming and Qian, Qi and Hasan, Kamrul and Du, Simon},
  title   = {Position: Universal Aesthetic Alignment Narrows Artistic Expression},
  journal = {arXiv preprint arXiv:2512.11883},
  year    = {2025}
}

@inproceedings{zhang2026genre,
  author    = {Zhang, Yi and Zhou, Wei and Zhao, Yang and Tan, Wenhao and
               Imoto, Keisuke and Gong, Zhen},
  title     = {Genre Bias or Aesthetic Perception? Identifying and Mitigating
               Shortcut Learning in Music Evaluation},
  booktitle = {International Society for Music Information Retrieval Conference
               (ISMIR)},
  year      = {2026},
  note      = {arXiv:2607.13903}
}

\end{document}